\documentclass[letterpaper, 10 pt, conference]{ieeeconf}

\usepackage{graphicx}
\graphicspath{ {./Figures/} }
\IEEEoverridecommandlockouts
\usepackage[UKenglish]{babel}
\usepackage[utf8]{inputenc}
\usepackage[T1]{fontenc}
\usepackage{newfloat}
\usepackage{booktabs}
\usepackage{amssymb, amsmath, array, bm, algorithm, standalone,csquotes,textcomp, amssymb,amsfonts, tabularx}
\usepackage{gensymb, wasysym} 
\usepackage{graphicx, xcolor, subcaption, soul}
\usepackage{siunitx, nameref, zref-xr}
\usepackage{algorithmic}
\usepackage{comment}
\usepackage[]{mdframed}
\zxrsetup{toltxlabel}
\usepackage{microtype}
\usepackage[nolist]{acronym}
\definecolor{myBlue}{HTML}{03456A} 
\definecolor{Gray}{RGB}{230,230,230}
\usepackage{hyperref}
\hypersetup{
    colorlinks=true,
    linkcolor=myBlue,
    citecolor=myBlue,
    filecolor=myBlue,      
    urlcolor=myBlue,
    pdftitle={main},
    pdfpagemode=FullScreen,
    }
\usepackage[nameinlink]{cleveref}
\crefname{figure}{Fig.}{Figs.}
\Crefname{figure}{Fig.}{Figs.}
\crefname{table}{Table}{Tables}
\Crefname{table}{Table}{Tables}
\usepackage{siunitx}

\newcommand\etal[0]{\textit{et\,al.\ }}
\newcommand\mailto[1]{\href{mailto:#1}{#1}}

\let\oldcite\cite
\renewcommand*\cite[1]{\,\oldcite{#1}}
\def\BibTeX{{\rm B\kern-.05em{\sc i\kern-.025em b}\kern-.08em
    T\kern-.1667em\lower.7ex\hbox{E}\kern-.125emX}}

\begin{document}
\begin{acronym}
    \acro{DMS}{Distributed Manipulator Systems}
    \acro{DoF}{Degrees of Freedom}
    \acro{PLA}{Polylactic acid}
    \acro{FOC}{Field Oriented Control}
    \acro{SPI}{Serial Peripheral Interface}
    \acro{PVC}{Polyvinyl chloride}
\end{acronym}

\title{Scalable Low-Density Distributed Manipulation Using an Interconnected Actuator Array\\
}

\author{
    Bailey Dacre,
    Rodrigo Moreno,
    Jørn Lambertsen,
    Kasper Stoy,
    Andrés Faíña\\
    Email:\mailto{\{baid, rodr, jrnl, ksty, anfv\}@itu.dk}
}

\maketitle

\begin{abstract}
Distributed Manipulator Systems, composed of arrays of robotic actuators necessitate dense actuator arrays to effectively manipulate small objects. 
This paper presents a system composed of modular 3-DoF robotic tiles interconnected by a compliant surface layer, forming a continuous, controllable manipulation surface. The compliant layer permits increased actuator spacing without compromising object manipulation capabilities, significantly reducing actuator density while maintaining robust control, even for smaller objects.
We characterize the coupled workspace of the array and develop a manipulation strategy capable of translating objects to arbitrary positions within an $N \times N$ array. The approach is validated experimentally using a minimal $2 \times 2$ prototype, demonstrating the successful manipulation of objects with varied shapes and sizes.
\end{abstract}


\section{Introduction}

Robotic manipulation often evokes images of biomimetic robot arms grasping objects. While such systems have seen significant adoption, they can face challenges when manipulating objects of different size and geometries without the use of specialized grippers. In contrast, non-prehensile approaches to manipulation, such as multi-actuator arrays, have demonstrated their utility in handling a variety of object morphologies. Arrays of actuators utilizing coordinated motion, \ac{DMS}, have been widely explored for their use in object manipulation tasks. These systems have been demonstrated from microscopic to macroscopic scales using diverse actuation mechanisms. Non-prehensile manipulation makes these systems well suited to handling soft objects or those with complex geometries. The use of many cooperating actuators provides these systems with high bandwidth, offering parallel manipulation capabilities and an inbuilt redundancy to failure. However, such systems are often restricted by the ratio of object size to actuator pitch, with most systems manipulating objects much larger than the inter-actuator spacing. As such, these systems are often actuator dense, which adds significant complexity both in control and maintenance.

\begin{figure}[t]
    \centering
    \includegraphics[width=\columnwidth, trim ={0 0 0  0}, clip]{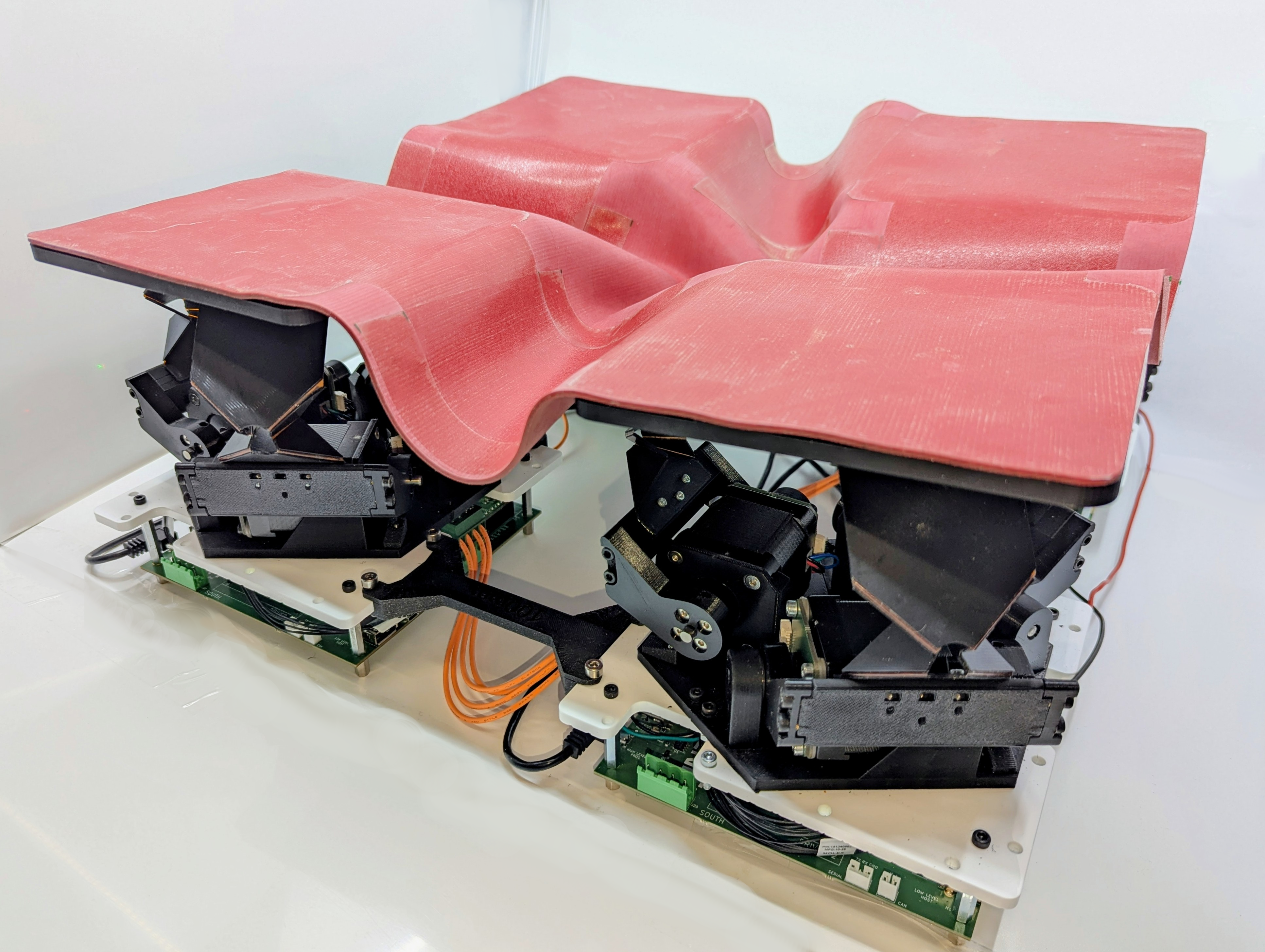}
    \caption{ A $2\times 2$ actuator array inter-connected by a flexible material to form a continuos manipulation surface}
    \label{fig:tile_array}
\end{figure}

In this paper, we present a \ac{DMS} composed of an array of robotic tiles interconnected by a compliant material to form a continuous manipulation surface. Each tile is a 3-\ac{DoF} parallel mechanism, with a flexible layer linking each end-effector. This layer ensures persistent object contact during manipulation, enabling increased actuator spacing without loss of control. Consequently, traditional object-to-actuator scale constraints are relaxed, allowing actuator density to be significantly reduced.

The main contributions of this work are:
\begin{itemize}
  \item Design and construction of a $2\times2$ array of 3-\ac{DoF} origami-inspired actuators, interconnected by a flexible layer to create a continuous manipulation surface
  \item Analysis of the workspace of the multi-manipulator array and the effects of inter-tile spacing and connective material length on achievable pose. 
  \item A state-machine-based controller enabling object translation to arbitrary positions, independent of array size.
  \item Experimental validation of manipulation capabilities for a wide variety of object geometries.
\end{itemize}

\section{Related Work}

\acp{DMS}, typically composed of two-dimensional actuator arrays, have been extensively studied at both the micro- and macro-scale for their effectiveness in object manipulation tasks. Owing to their spatial distribution, such systems can offer a high-throughput and parallel manipulation capabilities, making them advantageous for operations such as sorting and distribution. Various methods of actuation have been explored, including pistons \cite{follmerInFORMDynamicPhysical2013}, wheels\cite{uriarteControlStrategiesSmallscaled2019}, parallel-mechanism~\cite{patilLinearDeltaArrays2023, yimTwoApproachesDistributed2000}, and cilia-inspired MEMS actuators~\cite{bohringerTheoryManipulationControl1994}. These systems manipulate objects through the coordinated application of force from multiple actuators simultaneously and have demonstrated their ability to manipulate a wide range of objects with diverse geometries, depending on system configuration. 

Most \ac{DMS} share a common architectural characteristic: a dense actuator array manipulating objects substantially larger than the actuator pitch (the centre-to-centre spacing between adjacent actuators). This dense array ensures that objects maintain contact with multiple actuators simultaneously, providing stability and allowing for techniques such as gaiting. However, as multiple points of contact are required, performance degrades into inoperability as object size approaches or falls bellow the actuator pitch.

Control complexity increases proportionally to the number of \ac{DoF} of the system. Consequently, many systems utilizing dense arrays reduce control complexity by grouping actuators and driving them as a single unit~\cite{bohringerSensorlessManipulationUsing1995}. 

Manipulation at the scale of individual actuators has been investigated, such as the use of macroscopic three-\ac{DoF} cilia \cite{yimTwoApproachesDistributed2000}, capable of transporting spherical objects with single contact, though this required a dense array of actuators each with multiple degrees of freedom.

Actuated surfaces capable of altering their shape have been investigated for their potential in object manipulation tasks \cite{salernoOriPixelMultiDoFsOrigami2020, wangSurfacebasedManipulationModular2026, johnsonMultifunctionalSoftRobotic2023}, demonstrating versatile manipulation of a broad range of objects, including deformable or irregularly shaped objects.

The use of a connective material between actuators can create a continuous surface on which objects can be manipulated\cite{festoWaveHandling2013}. Such a surface ensures that objects remain in constant contact, regardless of their size. As pitch is no longer a critical restriction,  inter-connecting materials can enable a reduction in actuator density while retaining manipulation capabilities\cite{dacreFlexibleFoldableWorkspace2025, ingleSoftManipulationSurface2024}, although many such systems still rely on dense arrays and are often limited to rolling objects. 
This connectivity introduces opportunities for novel manipulation strategies, including those that leverage the mechanical properties of the compliant layer itself. This work extends prior demonstrations of linear arrays of interconnected 3\ac{DoF} actuators for object manipulation \cite{dacreFlexibleFoldableWorkspace2025} by generalizing the approach to a two-dimensional array, forming a continuous manipulation surface.

\section{Materials and Methods}

\subsection{Mechanical Design}

\begin{figure*}[t]
    \centering
    \includegraphics[width=0.85\textwidth]{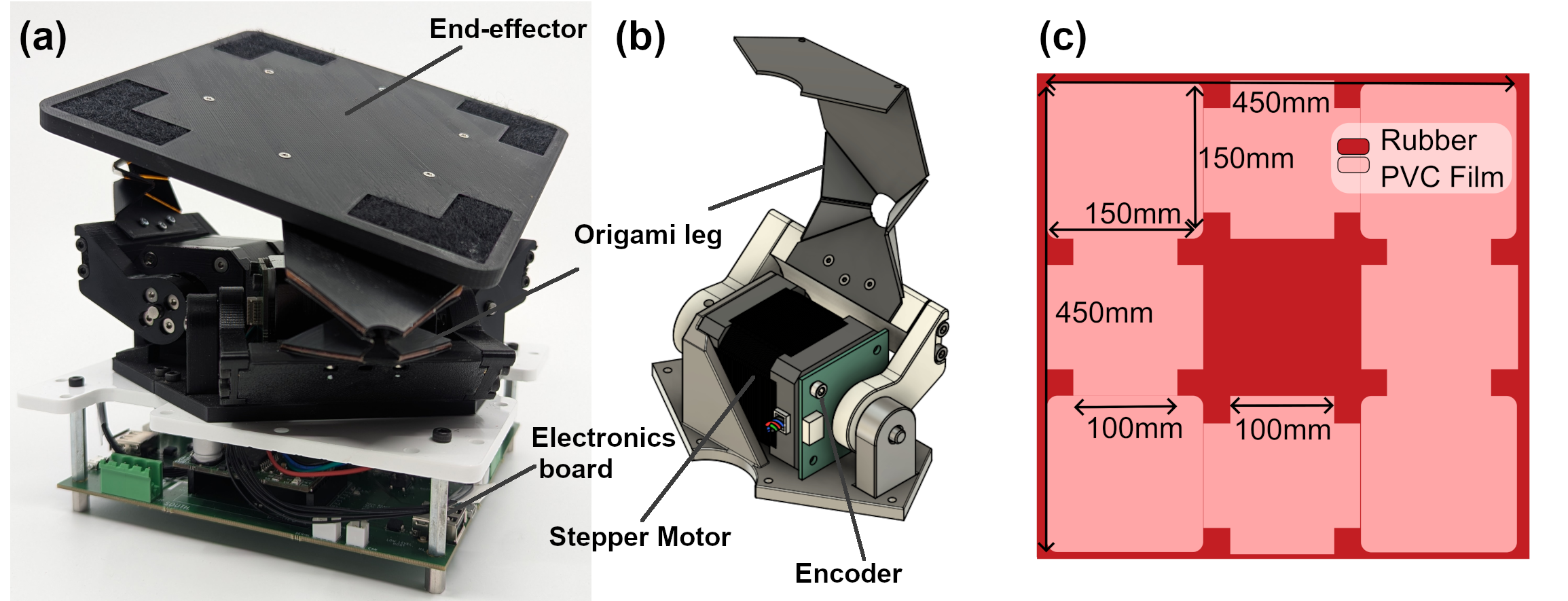}
    \caption{Array hardware: (a) Robotic tile module comprising three origami-inspired legs, each actuated by a stepper motor and connected to a central end-effector with attachment points for the flexible surface. (b) Detailed view of a single leg assembly. (c) Visualization of the compliant surface layer showing regions of exposed rubber and bonded PVC film.}
    \label{fig:hardware_collage}
\end{figure*}

The system comprises modular robotic tiles interconnected to form a continuous manipulation surface. Each of the modules is a 3-\ac{DoF} parallel mechanism, based upon the Canfield mechanism \cite{canfieldDevelopmentCarpalWrist1997}.  Each robotic module, referred to as a tile, features a square end-effector ($150mm \times 150mm$) supported by three foldable origami-inspired linkages, as can be seen in \cref{fig:hardware_collage}(a)(b). The origami linkages incorporate two revolute hinge joints and a central joint created using a waterbomb fold design, which allows the linkages to rotate about 2 axes, analogous to a spherical joint. 

The origami linkages follow the design shown in previous work \cite{dacreFlexibleFoldableWorkspace2025}. Each origami linkage consists of $0.15mm$ polyimide film embedded within $1.5mm$ PLA parts fabricated via fused deposition modelling. During fabrication, the print process is paused and the polyimide film is adhered in place. Holes in the polyamide allow for adhesion between the plastic layers, pinning the polyamide in place. Once complete, the polyimide remains partially exposed, forming flexible hinge lines while the surrounding PLA provides rigid structural segments.  Each linkage measures $150mm$ from the base to the revolute joint at the end-effector.

Each tile is actuated by three NEMMA 17 stepper motors (17HS19), with a holding torque of 59Ncm and a 1.8deg step angle. To track each motor's position, we utilize a 14-bit magnetic encoder (AS5048A), communicating over \ac{SPI}. \ac{FOC} was employed to position the motors, utilizing the SimpleFOC library \cite{skuricSimpleFOCFieldOriented2022}, enabling smooth torque and position regulation.

The tiles within the array are linked by a compliant interconnecting layer. This layer is a \SI{1.5}{\milli\meter}-thick Linatex\textregistered{} natural rubber. A \SI{0.1}{\milli\meter}-thick \ac{PVC} film is selectively bonded to the rubber using an acrylic adhesive. The addition of this film reduces surface friction, thereby improving the system’s object manipulation performance. However, the film also decreases flexibility in bonded regions. Consequently, its selective application, as illustrated in \cref{fig:hardware_collage}(c), enables a trade-off between friction reduction and surface compliance, optimizing performance for the intended manipulation tasks.

\subsection{Electronics, Communications, and Control}

 Each tile contains a Teensy\textregistered micromod microcontroller, which handles kinematic computations, motor control, and inter-module communication. Each motor is driven using a dual H-bridge motor driver (L6205PD), integrated into a custom PCB.

Each tile features five connectors interfaced with the Teensy’s serial ports. Commands can be sent from an external controller via USB or through the serial connections located on each of the tile’s four cardinal edges. These serial links allow the controller to communicate with neighbouring tiles in all directions and to forward commands. Power is distributed in a daisy-chain configuration, with each module capable of handling up to 30A of pass-through current, enabling up to eight tiles to be connected in series.

\subsection{System Configuration}

For this experimental setup, a system consisting of 4 tiles was connected in a $2\times2$ array. The tiles are evenly spaced, with centre-to-centre spacing, $D$, of $261\si{\milli\meter}$ between vertically and horizontally adjacent tiles. Tiles are then connected by a flexible connective layer to form a singular connected distributed \ac{DMS}. The material used for the inter-connective material and the inter-tile length, $L$, of $150\si{\milli\meter}$. Note, that due to the end-effector width, it is possible for $L\le D$.

\section{Kinematics and Workspace}
\subsection{Single Tile Kinematics}

\begin{figure}[t]
    \centering
    \includegraphics[width=\columnwidth, trim={20 0 400 0}, clip]{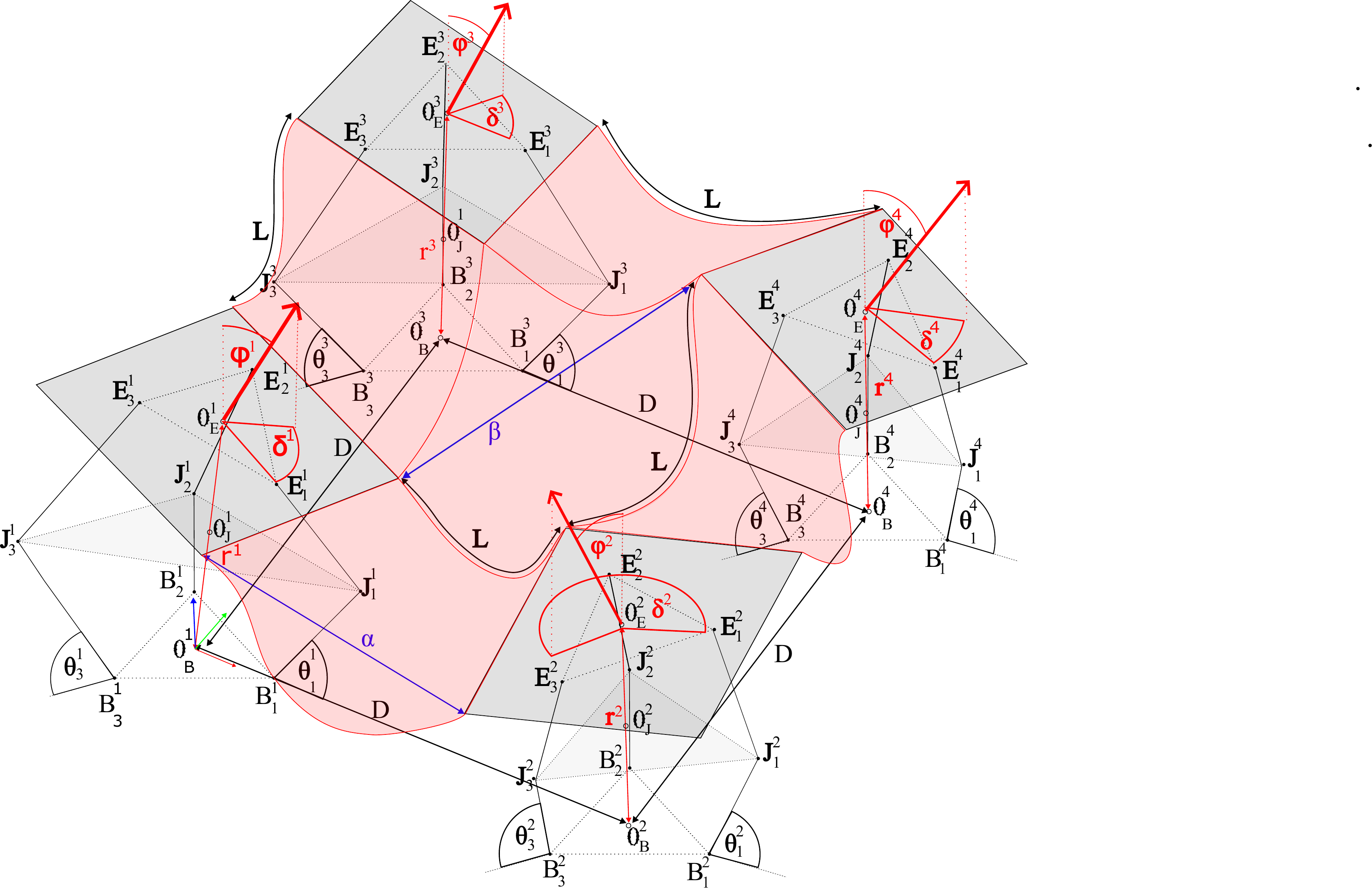}
    \caption{Illustration of four modules modules in an array, connected by connective material. This shows motor angles $\theta_i$, which determine tile pose $(\delta^i, \phi^i, r^{i})$. Modules are at an inter-module distance, $D$, and are connected by a material of length $L$, with  distances $\alpha$ and $\beta$ indicated. }
    \label{fig:four_tile_illustration}
\end{figure}

The system is an array of robotic tiles. To understand the kinematics and workspace of the system as a whole, we must first understand those of a single tile.

The kinematics of a parallel mechanism equivalent to that used by each tile has been previously analysed  \cite{meteClosedLoopPositionControl2021, dacreFlexibleFoldableWorkspace2025}.  The workspace of the parallel mechanism is determined by the geometric parameters of leg length and leg position. The length of the origami legs, $l$, is divided  equally by the central waterbomb joint. The base of each leg sits equidistant from the centre of the robots base, residing on a circle of radius $R$. In our system, leg length $l = \SI{140}{\milli\meter}$, $R = \SI{44.01}{\milli\meter}$, and the base of the legs are positioned at $\Phi_1 = \frac{\pi}{3}$, $\Phi_2 = \pi$, $\Phi_3 = \frac{5\pi}{3}$, where $\Phi_i$ is the angle about the z axis of the local coordinate system with origin at the centre of the robots base, $O_B$.

The position of the end-effector centre, $O_E$, is determined by the angles of the legs, $\theta_i$,  with respect to the $XY$ plane. The values of $\theta_i$, are constrained to the range $0\le \theta_i\le  \frac{7\pi}{18}$, a restriction on the achievable $\frac{\pi}{2}$, implemented to protect the system from damage due to hyper-extension . The position of $O_E$ is expressed in polar coordinates $(\delta, \phi, r)$, where $r =||\overrightarrow{O_BO_E}||$, $\delta$ is the angle of rotation about the Z axis (yaw), and $\phi$ is the angle of rotation about the transformed Y axis (composite pitch and roll), which determines end-effector tilt. A visualization of system angles can be seen in \cref{fig:four_tile_illustration}. 

\begin{figure}[t]
    \centering
    \includegraphics[width=0.7\columnwidth]{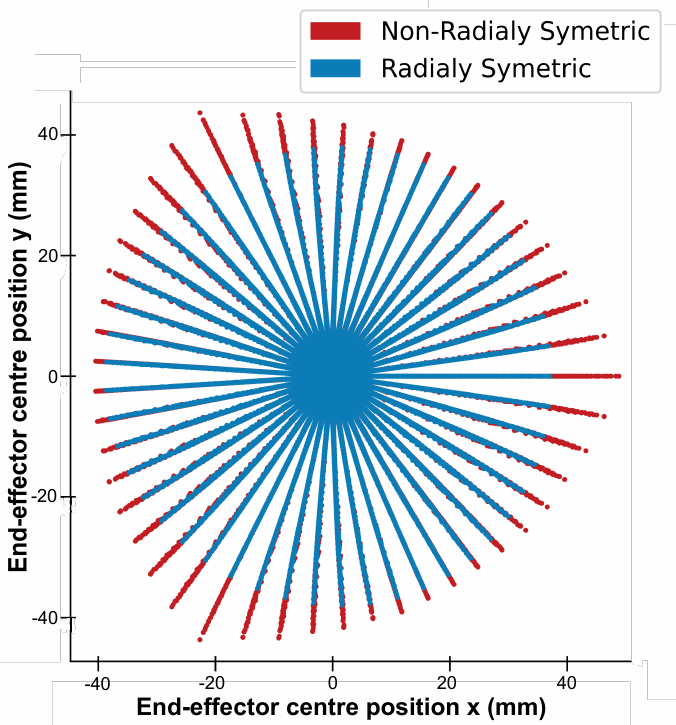}
    \caption{Visualization of the workspace of a single tile in red, and a radially symmetric subsection of that workspace in blue. Points show position of end-effector centre.}
    \label{fig:workspace_comparison}
\end{figure}

\Cref{fig:workspace_comparison} shows a visualization of a single tile's workspace from above. The visualization is created by scattering the calculated end-effector positions for valid solutions to the systems inverse kinematics by sweeping through values of $\delta$, $\phi$, and $r$.

\subsection{Connected Tile Constraints}

The system comprises a two-dimensional array of tiles whose end-effectors are linked by a compliant material. While any tile can individually reach any point in its workspace, restrictions are imposed on tile poses that can be reached simultaneously in the connected array without straining the connective material. If we consider this to be undesirable, such as to avoid excessive motor load, we have a restriction on the arrays shared workspace. Such a workspace restriction is determined by the relative positions of the tiles, the dimensions of the end-effector, and the length of material that connects each tiles actuator, $L$. In this work, we explore a system in which tiles are laid out in a regular grid pattern, with a distance between the base of horizontally or vertically adjacent tiles being $D$. 

For a tile in an arbitrary sized array, if a tile is not on a boundary, we must consider the the eight neighbouring tiles when determining if all poses can be achieved simultaneously. By using the relative positions of the corners of two adjacent tile's square end-effectors, along with the known length of the material connecting them, it is possible to determine whether a given pair of simultaneous tile poses would cause the connecting material to exceed its unstrained length.

As described by Meta \etal \cite{meteClosedLoopPositionControl2021}, the translation vector, $t$, between the base, $O_B$, and the centre of the end-effector, $O_E$, of such a system is given by:

\begin{equation}
t = \overrightarrow{O_BO_E} = r 
\begin{bmatrix}
\sin \phi \cos \delta \\
\sin \phi \sin \delta \\
\cos \phi
\end{bmatrix}
= r \vec{n}
\end{equation}

where $\vec{n}$ is a unit vector in the direction of $O_E$ and r is the distance between $O_B$ and $O_E$.

A local coordinate frame is defined at point $O_E$, aligned with the end-effector’s orientation. The rotation between the base and end-effector frames is given by the extrinsic composition of rotations about the base $z$- and $y$-axes:

\begin{equation}
R_z(\delta)=
\begin{bmatrix}
\cos\delta & -\sin\delta & 0\\
\sin\delta &  \cos\delta & 0\\
0 & 0 & 1
\end{bmatrix},
R_y(\phi)=
\begin{bmatrix}
\cos\phi & 0 & \sin\phi\\
0 & 1 & 0\\
-\sin\phi & 0 & \cos\phi
\end{bmatrix}
\end{equation}

\begin{equation}
R = R_z(\delta) R_y(\phi).
\end{equation}

The homogeneous transformation between the two frames is then defined as:

\begin{equation}
H =
\begin{bmatrix}
R & \mathbf{t}\\
\mathbf{0}^\top & 1
\end{bmatrix}
\in \mathbb{R}^{4\times4}.
\label{eq:homogeneous}
\end{equation}

The positions of the corners, $C$, of a square end-effector with edge length $E_W$, and height $E_H$ are given in the end-effector coordinate system by:

\begin{equation}
\mathcal{C}
=
\left\{
\begin{bmatrix}
i \tfrac{E_W}{2},
j \tfrac{E_W}{2},
E_H
\end{bmatrix}^\top
:\;
i,j \in \{+1,-1\}
\right\}
\label{eq:local_corners_set}
\end{equation}

The relative positions of neighbouring tile corners are used to assess whether a given pair of poses would strain the connecting material. Horizontally and vertically adjacent tiles are assumed to be joined at their nearest edges, with the distal corners of those edges defining the maximum separation $\alpha$. Diagonally adjacent tiles are considered connected at their nearest corners, defining the separation $\beta$, as illustrated in \cref{fig:four_tile_illustration}. Adjacent end-effectors are linked by material of length $L$, and for a uniform grid, the material length connecting diagonally adjacent tiles is $\sqrt{2}L$. We then impose the following constraints: $L \ge \alpha$ and $\sqrt{2}L \ge \beta$. A tile pose is considered valid only if, for all neighbouring tiles, these constraints are not violated.

\subsection{Shared Workspace}

For a set of poses to exist within the shared workspace of the array, all adjacent tiles must simultaneously satisfy geometric constraints. In a discretized workspace with \(P\) poses per tile, a naïve approach would require evaluating all pose combinations for a tile and its up to eight neighbours, resulting in \(kP^9\) computations.

This cost can be reduced by exploiting symmetries in the array. In a uniform grid, connections between a central tile and its neighbours are equivalent in magnitude to those between neighbouring tiles, allowing redundant evaluations to be removed. Once the validity of a pose pair is established for one pair of tiles, all equivalent connections are implicitly satisfied. This allows sequential rather than combinatorial evaluation, significantly reducing the total computational effort.

A square array exhibits four lines of symmetry, while each tile’s workspace has three, as seen in \cref{fig:workspace_comparison}, owing to its leg geometry. When these symmetries are aligned, many connections become equivalent under symmetry transformations. Thus, for the central tile, it is enough that all $P$ positions within its workspace are compared against the full workspace of three neighbouring tiles and half of the workspace of two others, whose domains are bisected by a symmetry line. This reduces the total required evaluations to approximately $4kP^2$.

Larger tilt angles, \(\phi\), occur when \(\delta\) aligns with one of the tile’s legs. By constraining the single-tile workspace to be radially symmetric—so that the maximum reachable \(\phi\) at each radius \(R_0\) is independent of \(\delta\) (illustrated in blue in \cref{fig:workspace_comparison})— the number of symmetries in the array increases to three. These symmetries significantly reduce redundant configurations, reducing the computational cost of determining the shared workspace to only $kP^2$ evaluations.

\begin{figure}[t]
    \centering
    \includegraphics[width=\columnwidth, trim={0 0 0 0}, clip]{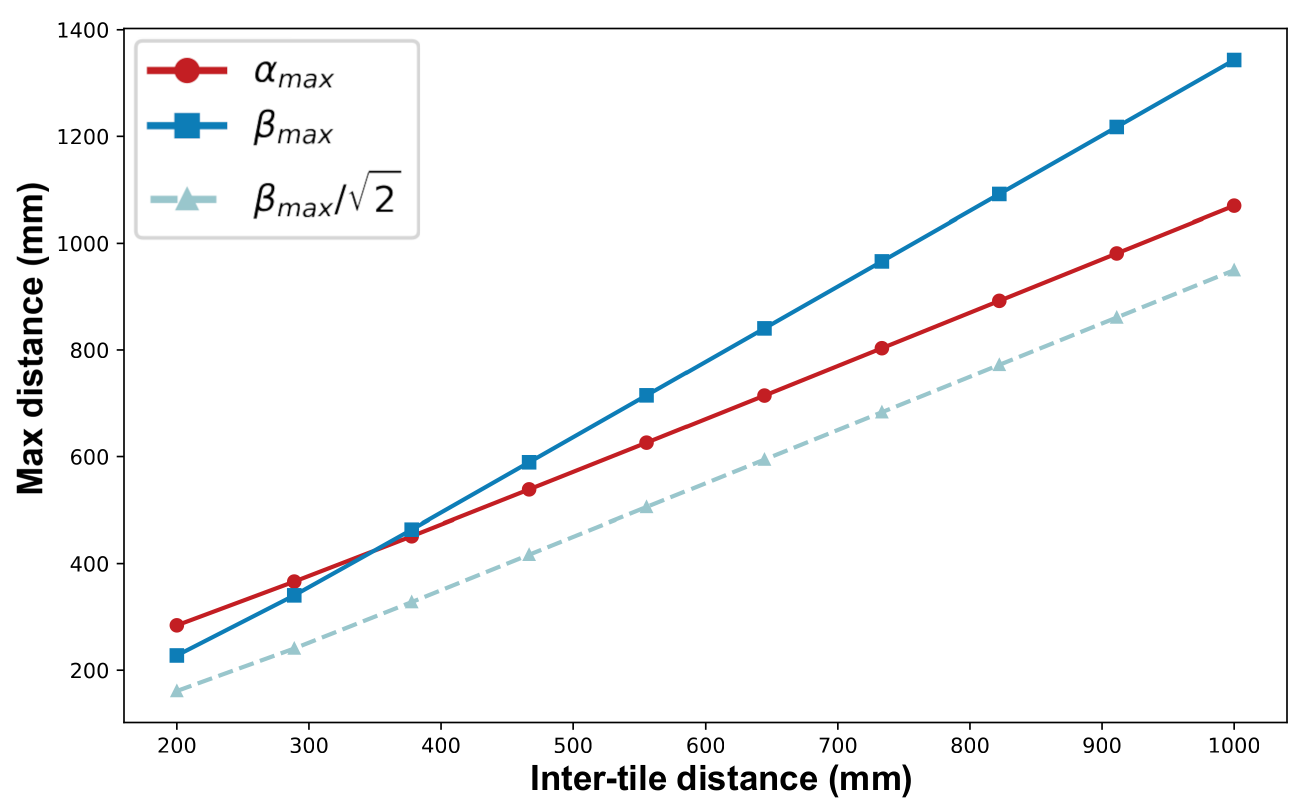}
    \caption{Minimum material length to achieve all positions in single tile workspace when connected to a horizontally/ vertically adjacent tile ($\alpha_{max}$) and diagonally adjacent tile ($\beta_{max}$) for different inter-tile distances (D). }
    \label{fig:plot_alpha_beta_distance}
\end{figure}

Using the radially symmetric workspace, the minimum interconnecting material length \(L_{\min}^D\) can be efficiently determined for any inter-tile spacing \(D\), informed by the maximum values of $\alpha$ and $\beta$, $\alpha_{max},\beta_{max}$. \Cref{fig:plot_alpha_beta_distance} shows their variation with \(D\), defining a lower bound on \(L\) that ensures neighbouring tile workspaces remain decoupled. The values of \(\alpha_{\max}\) and \(\beta_{\max}/\sqrt{2}\) scale proportionally with \(D\), with \(\alpha_{\max}\) consistently larger. Hence, for regular square arrays, satisfying \(L \ge \alpha_{\max}\) is sufficient to guarantee decoupling, eliminating the need to compute \(\beta_{\max}\).

\subsection{Pulling Taut and Maximizing Material Tilt}

As each tile rotates about its base rather than the centre of its end-effector, the inter-tile material slackens when neighbouring tiles tilt toward each other and tightens when they tilt apart. As demonstrated previously \etal~\cite{dacreFlexibleFoldableWorkspace2025}, this behaviour can be exploited for object manipulation: objects are captured by regions of slack and then pushed out as the material becomes taut. This strategy is used to transition objects from the inter-tile region onto the tile surface.

For a given inter-tile spacing \(D\) and material length \(L\), workspace analysis enables positioning of a neighbouring pair of tiles - the assisting and receiving tiles - such that their poses satisfy \(\alpha = L\), pulling the material taut.  

\begin{figure}[t]
    \centering
    \includegraphics[width=\columnwidth, trim={0 0 0 0}, clip]{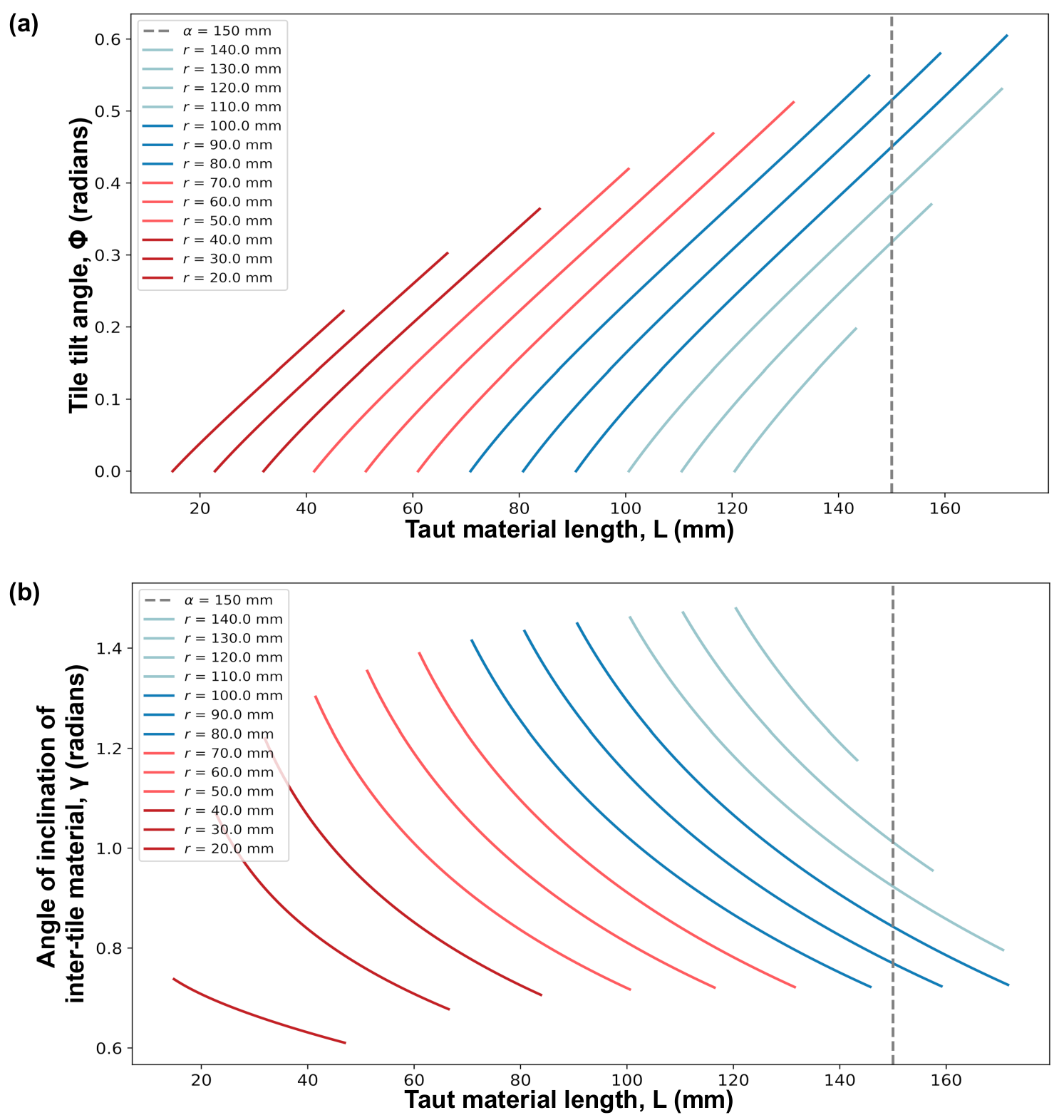}
    \caption{Effect of inter-tile material length on tile pose. One tile remains fixed while the other is positioned at \(D = \SI{261}{\milli\meter}\) with \(\delta\) directed away. (a) Maximum reachable \(\phi\) and (b) maximum \(\gamma\) for different \(r\) values. The grey line indicates the material length used in the physical prototype.}
    \label{fig:double_plot_L_phi_gamma_r}
\end{figure}

To push an object toward the receiving tile, the assisting tile’s \(\delta\) should align with the vector connecting the tile centres. With the receiving tile at a fixed pose, \(\phi\) and \(r\) can then be adjusted to achieve \(\alpha = L\), as shown in \cref{fig:double_plot_L_phi_gamma_r}(a). Maximizing the inclination angle of the taut inter-tile material, \(\gamma\), promotes object motion. Its value can be derived from the location of the end-effector corners. \Cref{fig:double_plot_L_phi_gamma_r}(b) illustrates the relationship between \(\gamma\), \(\alpha\), and \(r\).  

Thus, for a given \(L\), a combination of \(r\) and \(\phi\) can be found that which simultaneously pulls the material taut and maximizes \(\gamma\), ensuring efficient object transfer.

\section{Control}

\subsection{Regions and Path Planning}

The manipulation surface is partitioned into three distinct types of region: tile regions positioned above an end-effector, the inter-tile material connecting adjacent tiles, and the central region between four neighbouring tiles. An array of arbitrary size can be segmented in the XY plane by defining boundaries according to the end-effector width, $E_w$, and the inter-tile spacing, $D$, resulting in clearly defined regions that correspond to each functional area of the surface.

The system is capable of manipulating objects across all regions and positioning them at any location atop the tile end-effectors. Path planning is performed based on the current and target positions of an object to determine an optimal traversal route across the array. A graph of connected regions is constructed based on adjacency, and Dijkstra’s algorithm is used to compute efficient paths between them. The plan is continuously updated whenever an object transitions between regions, automatically adapting if an object moves to an unexpected region, thereby enhancing robustness to unanticipated object motion. By default, connection weights are set according to the centre-to-centre distance between regions; however, these weights can be adjusted to favour or discourage specific transitions. For instance, moving an object from the central region to a tile region was found to be more challenging than the reverse, and the connection weights can be modified to discourage this transition.

\subsection{Controller}

\begin{figure}[t]
    \centering
    \includegraphics[width=\columnwidth, trim={0 30 20 30}, clip]{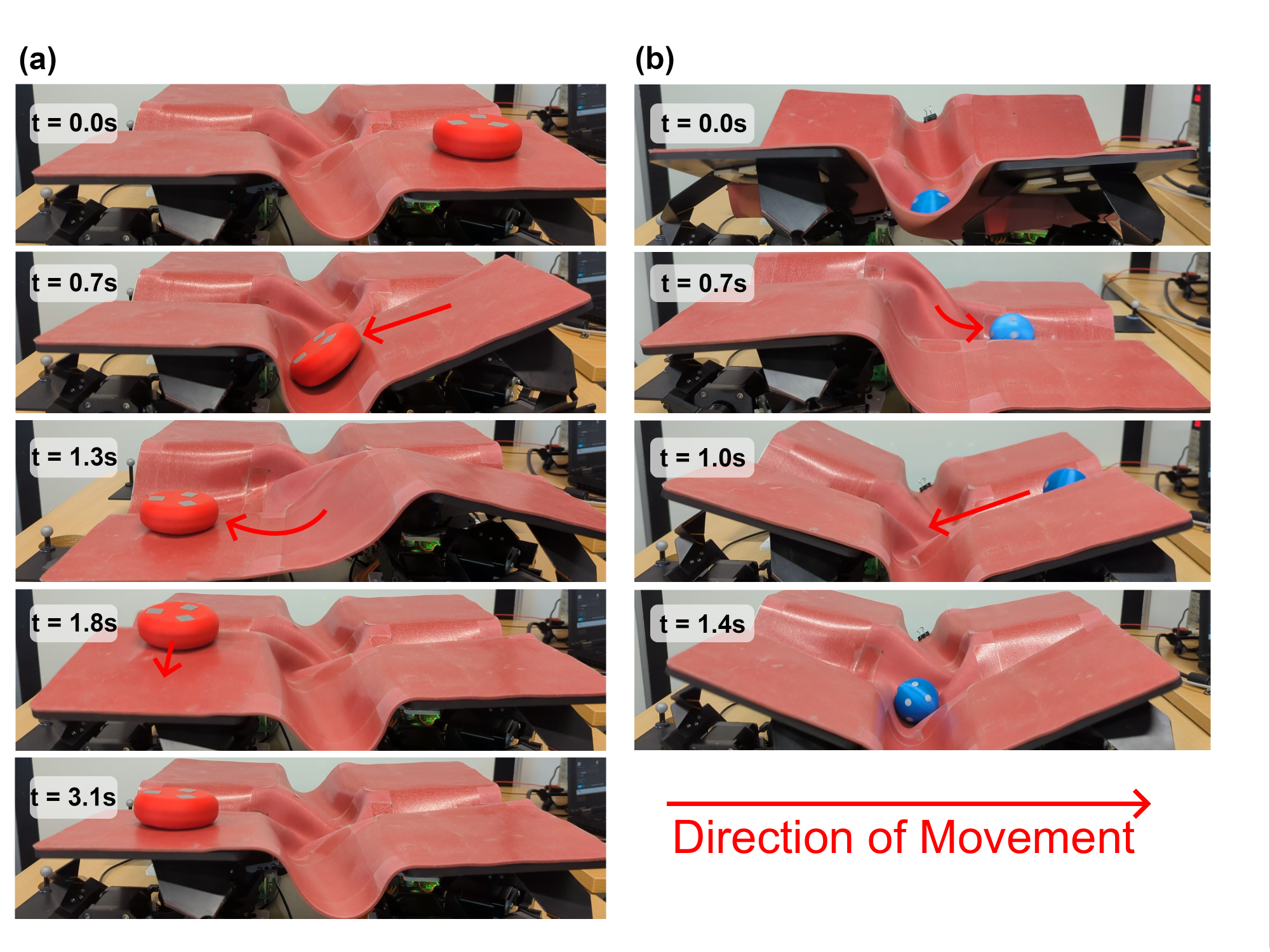}
    \caption{Object translation on the actuator array: (a) puck translating between tiles via coordinated motion; (b) sphere translating within the inter-tile compliant material.}
    \label{fig:object_translation_mosiac}
\end{figure}

The system controller is implemented as a state machine, where the current state is defined by: the object’s position, the next region in the planned path, the elapsed time within the current region, the region's type (tile end-effector or inter-tile connective material), and if the tile is currently moving.

The state governs the target pose of all tiles and whether each tile is actively moving toward it, holding position, or vibrating about this position.

A canonical list of poses used in the controllers state machine is listed in \cref{tab:tile_canonical}. These have a rotational symmetry; the poses for a specific state can be found by applying a rotation to each of these poses.

\begin{table*}[t]
\centering
\caption{Canonical states determining tile pose from current and target regions, listed in the form $(\delta, \phi, r)$. 
All other transitions are obtained by $R(\frac{\pi}{2}k)$ rotation symmetry, 
$k \in \{0,1,2,3\}$. For example, \(k = 0\) corresponds to the tile located in the north-west position of the array.  Angles are in radians, heights in millimetres.} 
\label{tab:tile_canonical}
\begin{tabular}{lcccc}
\toprule
\textbf{Current Region $\rightarrow$ Next Region} & \textbf{Tile NW} & \textbf{Tile NE} & \textbf{Tile SW} & \textbf{Tile SE} \\
\midrule
TILE $\rightarrow$ INTER\_TILE &
$(0, \frac{5\pi}{36}, 90)$ & (0, 0, 90) & (0, 0, 90) & (0, 0, 90) \\

INTER\_TILE $\rightarrow$ TILE &
$(0,0,10)$ & $(0,\frac{5\pi}{36},90)$ & (0, 0, 90) & (0, 0, 90) \\

INTER\_TILE $\rightarrow$ CENTRE &
$(\frac{3\pi}{2},\frac{5\pi}{36},90)$ & $(\frac{3\pi}{2},\frac{5\pi}{36},90)$ & $(\frac{\pi}{2},\frac{5\pi}{36},90)$ & $(\frac{\pi}{2},\frac{5\pi}{36},90)$ \\

CENTRE $\rightarrow$ INTER\_TILE &
$(0,0,10)$ & $(\pi,0,10)$ & $(\frac{5\pi}{4},\frac{\pi} {12},90)$ & $(\frac{7\pi}{4},\frac{\pi} {12},90)$ \\
\bottomrule
\end{tabular}
\end{table*}

To prevent propulsive motion, when moving to a new pose, all tiles follow a trajectory plan that is informed by a maximum velocity, $vel_{max}$, and acceleration, $acc_{max}$, in motor angle space. Here we use $vel_{max} = \frac{20 \pi}{9} rad \cdot s^{-1} $  and $acc_{max} = \frac{5 \pi}{6} rad \cdot s^{-2}$.

When entering a region that contains a tile-end-effector, the object will be centralized within the region before continuing. This was found to increase translation repeatability by ensuring a consistent starting position. A PID controller was implemented to adjust tile $\phi$ to tilt objects towards the centre, with $\delta$ set by taking the relative angle between the object and centre.

Occasionally, objects can become stuck when translating between regions. Therefore, if an object has been within a region for more than 5 seconds, all tiles will begin to oscillate about their previous pose with an additional sinusoidal adjustment made to the value of $r$. The amplitude of this oscillation was set to $10 \si{\milli\meter}$ with a frequency of $10\si{\hertz}$

A visualization of inter-tile and intra-tile object manipulation is shown in \cref{fig:object_translation_mosiac}, illustrating object positioning and translation between regions. In the first sequence (puck, Fig.~\ref{fig:object_translation_mosiac}a), the object is initially centred on a tile, then displaced into the inter-tile region. Coordinated motion of the adjacent tiles subsequently pushes the object onto the neighbouring tile, where it is re-centred in preparation for further translation. In the second sequence (ball, Fig.~\ref{fig:object_translation_mosiac}b), the object translates entirely within the inter-tile material. Local deformation of the compliant layer drives the object outward, after which it returns toward the central region by the collaborative tilt of the two adjoining tiles.

\section{Object Manipulation Experiments}

To demonstrate the utility of the system to perform object translation, we performed a number of experiments on a small scale $2\times2$ tile array.

\subsection{Experimental Setup}

A variety of objects were tested to verify the manipulation capabilities of the system across different geometries and sizes. This included: a cube ($\SI{40}{\milli\meter}^3$ ), a puck ($\diameter \SI{70}{\milli\meter} \times \SI{25}{\milli\meter}$), a sphere ($r = \SI{25}{\milli\meter}$), and a small tetrahedron with rounded corners(side length $=\SI{20}{\milli\meter}$).

Object position was determined using a 6-camera OptiTrack marker based motion capture system. 

Objects were considered to be within a region if their centre remained inside the region for at least 0.5s, which mitigates rapid state transitions caused by transient motion or sensor noise. For positioning or centring tasks, an object was considered to have reached a target point if its centre stayed within $\SI{15}{\milli\meter}$ of the target for at least 1s.

\subsection{Cyclic translation}

\begin{figure}[t]
    \centering
    \includegraphics[width=0.85\columnwidth, trim={0 0 0 0}, clip]{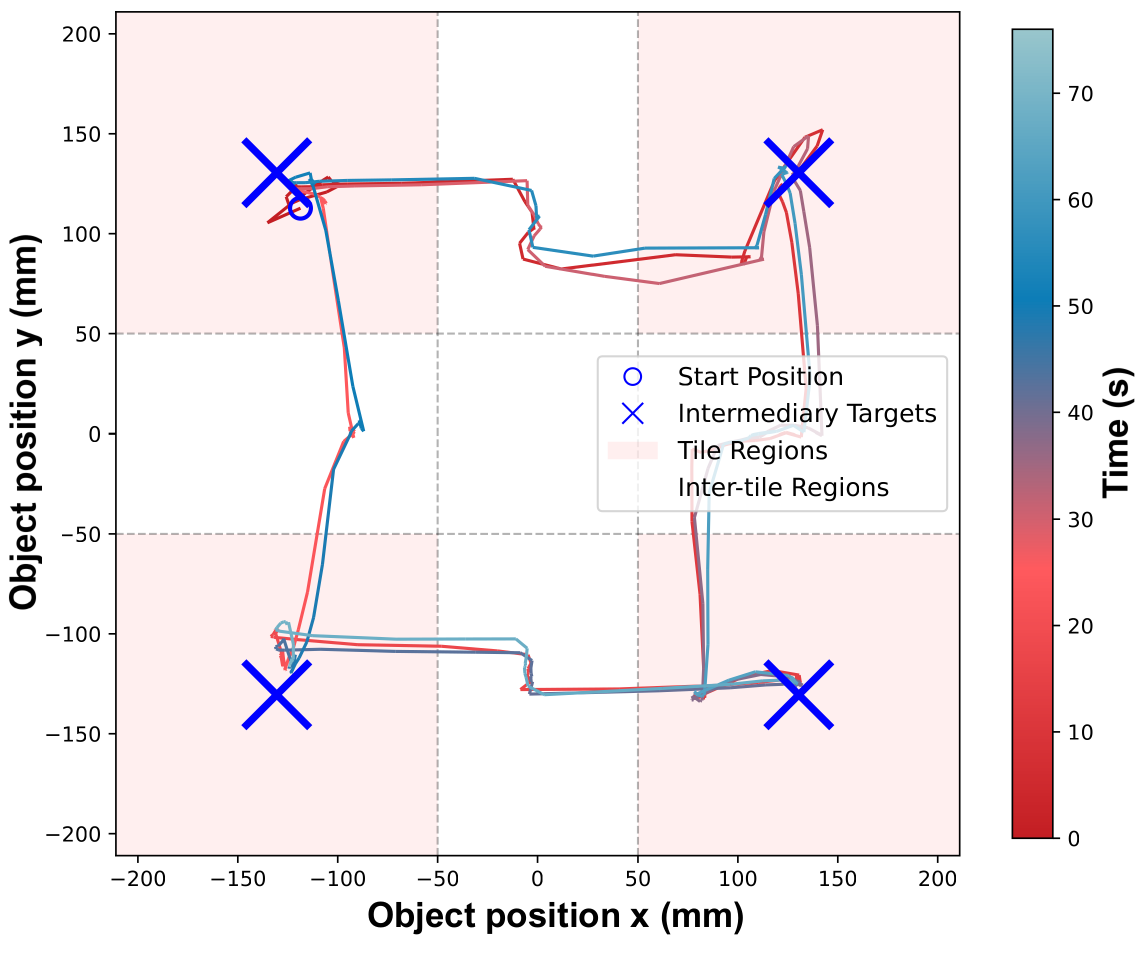}
    \caption{Trajectory of puck performing three cyclic translation around the array. Intermediary targets highlighted.}
    \label{fig:cycle_puck}
\end{figure}

\begin{figure}[t]
    \centering
    \includegraphics[width=0.85\columnwidth, trim={0 0 0 0}, clip]{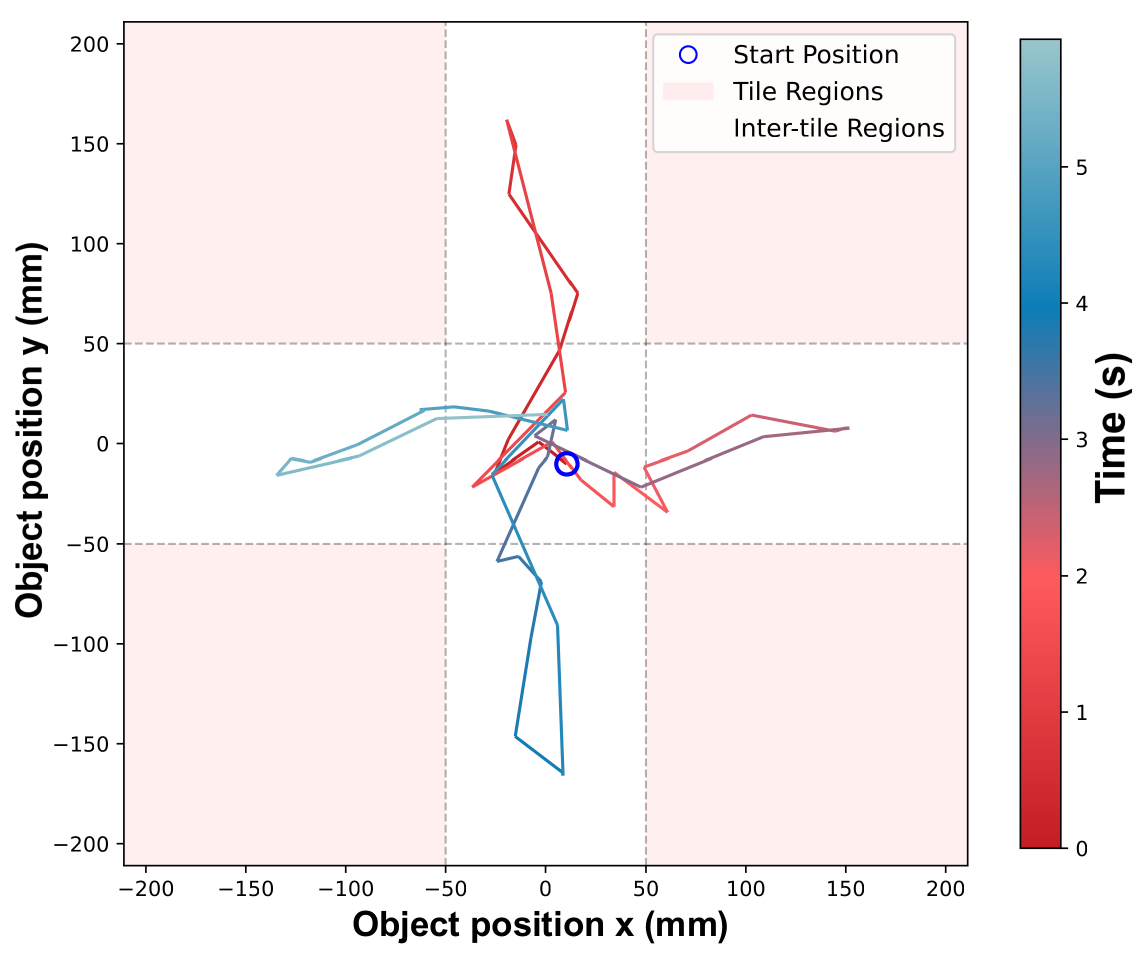}
    \caption{trajectory of sphere performing cyclic translation between inter-tile regions. No distinct targets positions, as here targets are just region transitions.}
    \label{fig:cycle_sphere}
\end{figure}

Objects manipulated on the array can translate through two distinct modes of transport. The first is tile-to-tile transfer, in which the object moves across the array by sequentially transitioning between adjacent tile regions via the inter-tile material. The second mode consists of continuous translation through the inter-tile material without direct alignment over any single tile. \Cref{fig:object_translation_mosiac} shows a visual representation of these motions.

To demonstrate both transport mechanisms, cyclic manipulation experiments were performed. \Cref{fig:cycle_puck} and \cref{fig:cycle_sphere} show object trajectories in which objects were translated along closed-loop trajectories, either between tiles or entirely through the inter-tile material. 

\begin{figure}[t]
    \centering
    \includegraphics[width=0.72\columnwidth, trim={0 0 0 0}, clip]{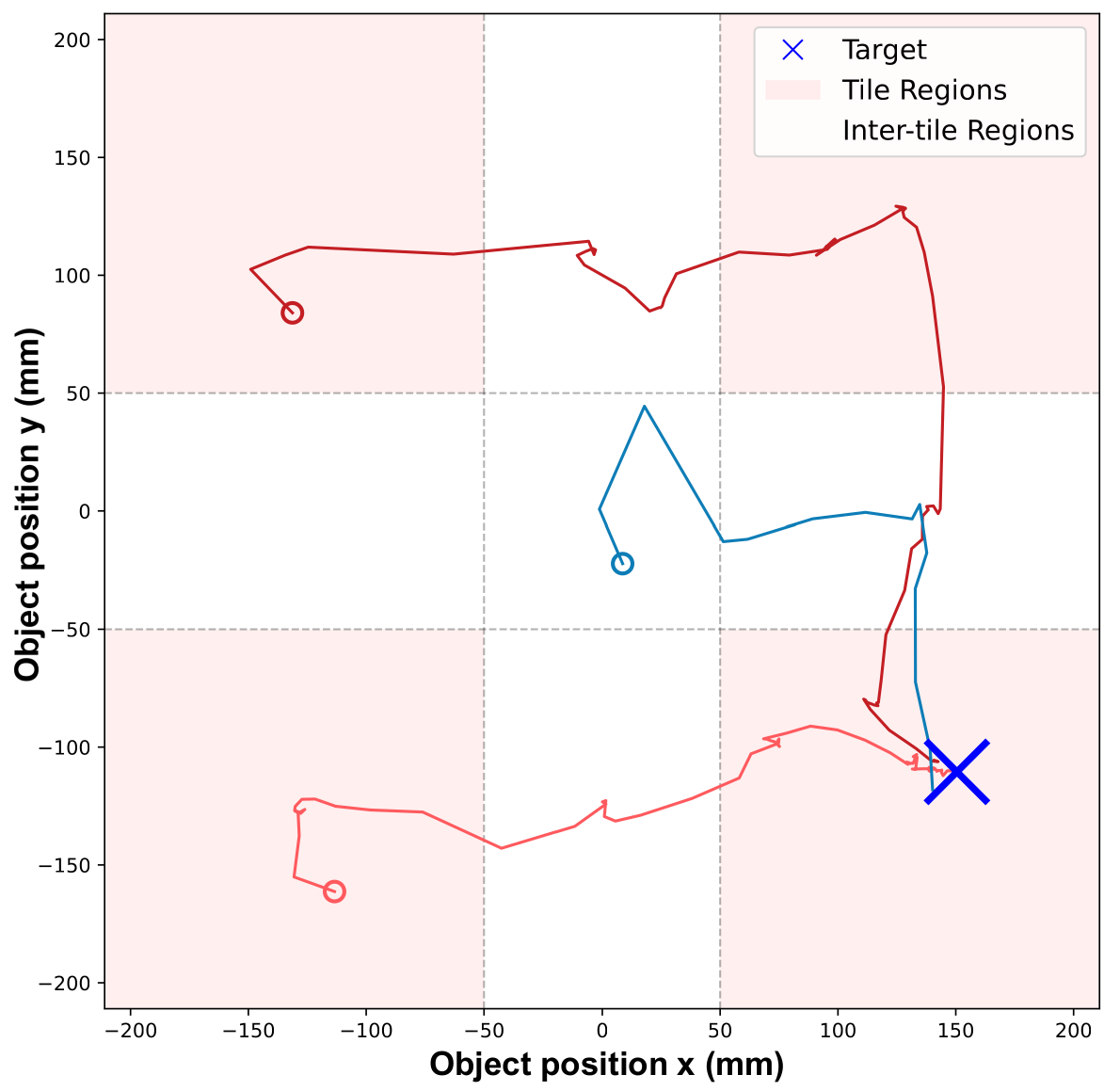}
    \caption{Trajectories of cube moving towards a an arbitrary target (150.5mm,-110.5mm) positioned on a tile, translating  from multiple starting positions.}
    \label{fig:point_to_point_cube}
\end{figure}

\begin{figure}[t]
    \centering
    \includegraphics[width=0.72\columnwidth, trim={0 0 0 0}, clip]{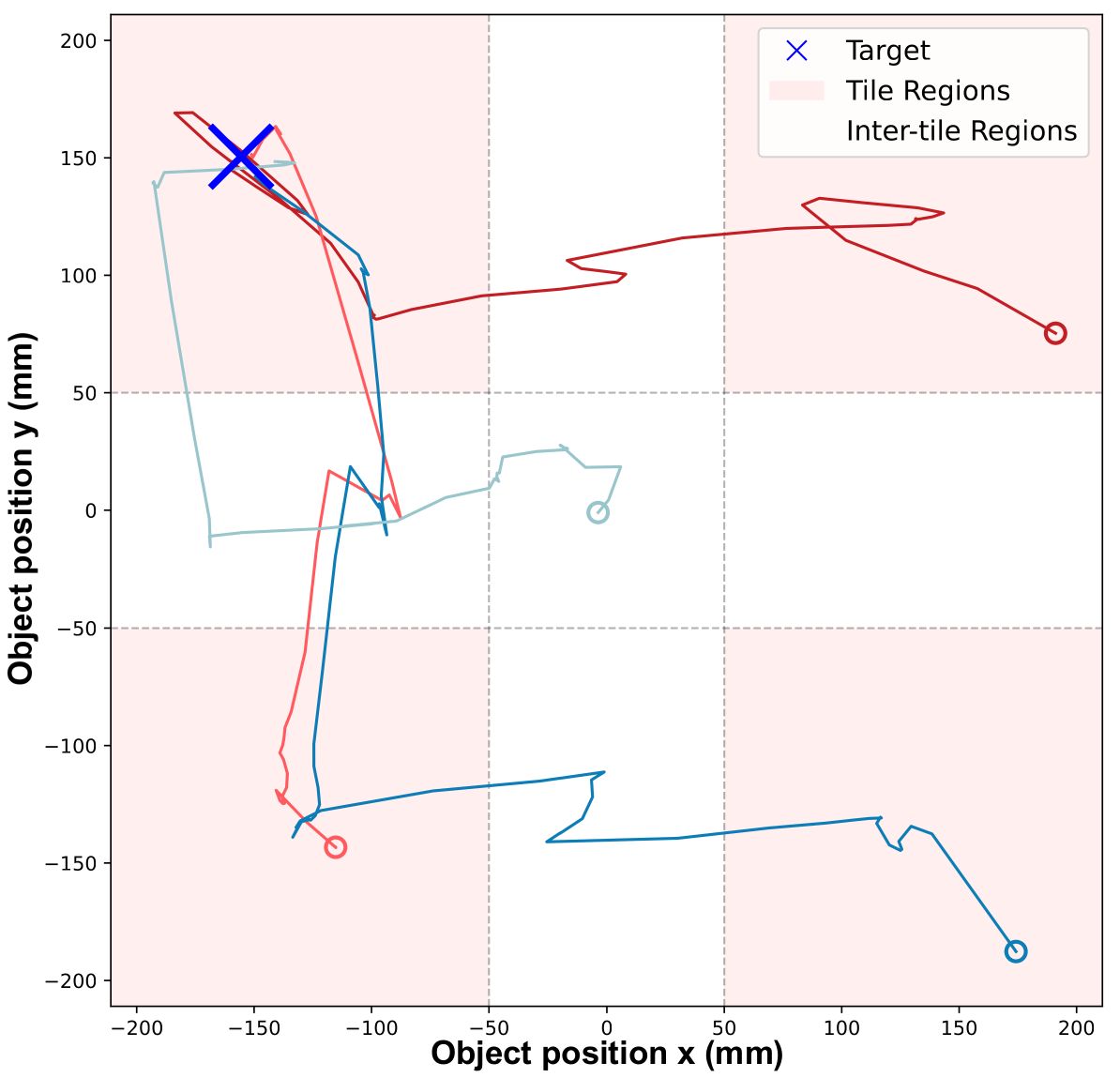}
    \caption{Trajectories of the tetrahedron moving towards a an arbitrary target (-155.5mm, 115.5mm) positioned on a tile, translating  from multiple starting positions.}
    \label{fig:point_to_point_pyramid}
\end{figure}

\subsection{Point to Point translation}

The system is capable of translating objects to arbitrary positions on each tile's end-effector. To validate this, target points were arbitrarily selected across the array and objects were translated to these locations from a variety of initial positions. As demonstrated in \cref{fig:point_to_point_cube} and \cref{fig:point_to_point_pyramid} by the cube and tetrahedron, objects can be successfully manipulated to target locations from multiple diverse starting locations.

Point positioning of objects within the inter-tile regions was not performed in this work. Although such positioning may be feasible, stable equilibrium points do not exist in these areas, preventing the system from positioning and maintaining an object at an arbitrary location and returning to a neutral state once positioned.

\section{Discussion}

In this work, the manipulated objects are smaller than the tile end-effectors and inter-tile regions. Therefore, each object interacts with at most four tiles at once. When tiles move independently (\(L \ge L_{\max}^D\)), manipulation can be controlled purely by considering this  locally four-tile region. This locality suggests that this same control principles can scale to larger \(N \times N\) arrays. 
The architecture therefore supports scalable manipulation through spatially distributed, locally defined rules, without requiring centralized global surface planning.

The hybrid soft–rigid design has several advantages: rigid end-effectors and linkages provide predictable kinematics and force transmission, while the compliant inter-tile layer ensures continuous contact across tile boundaries. When unconstrained by neighbours, actuators are largely decoupled, reducing manipulation to a local coordination problem while eliminating contact discontinuities and improving manipulation robustness across a variety of object geometries.

This work focuses on objects smaller than the actuator pitch; manipulating larger objects would require alternative strategies, such as gated or sequential actuation, and simultaneous consideration of a larger set of tiles~\cite{salernoOriPixelMultiDoFsOrigami2020}.

Two distinct modes of translation are demonstrated: tile-to-tile motion and via the connective material. All tested objects were capable of translation via both methods. However, sliding objects, such as the puck, exhibited more reliable motion using tile-to-tile translation, with reduced instances of becoming stuck on the material. Conversely, rolling objects, such as the sphere, translated more efficiently through the inter-tile material, moving rapidly between regions with lower risk of losing control than on the tile surfaces. These results suggest that the optimal manipulation strategy is dependent on object geometry and contact dynamics.

A distinctive feature of the \(2\times2\) actuator array, compared to a general \(N\times N\) configuration, is that all tiles exist on a boundary condition; no tiles are connected on all edges. For systems in which tiles are unconstrained by neighbours, their behaviour is largely comparable. However, the presence of boundary conditions introduces asymmetrical loading on edge tiles due to the weight of the interconnecting material. This effect can be observed in \cref{fig:cycle_puck}, where the manipulated object consistently drifted toward the centre of the array while traversing inter-tile regions, due to uneven loading. Future work will aim to experimentally validate the behaviour of larger arrays which contain fully interconnected tiles to confirm the scalability of the proposed architecture.

During workspace analysis and manipulation, the interconnecting material is modelled as inextensible; this does not hold for real-world materials. The approximation fails most notably in the central region of the array, where the material is required to bend simultaneously in two perpendicular directions as the tiles move. It is geometrically impossible for a continuous surface to maintain curvature in both directions without stretching.
Due to the relatively high elastic modulus of the rubber (\(\sim 2\,\text{MPa}\)), the material transitions between stable configurations through snap-through buckling when subjected to significant curvature. This non-linear deformation can result in unpredictable motion in the central region if not accounted for. The effect is amplified by the bonded acrylic film, whose stiffness constrains in-plane stretching. Therefore, the final design incorporates both bonded and un-bonded regions to allow for some compliance. Material compliance also introduces hysteresis and associated energy dissipation. 
Future work should quantify the influence of material stretch on object manipulation, reachable workspace, positional drift over repeated cycles, and motor loading, and investigate strategies to exploit or mitigate compliance-driven effects, including the use of materials with tailored anisotropic properties to reduce snap-through behaviour.

Another limitation of the current system lies in the use of static regions in the \(XY\)-plane to represent tile positions. In practice, each tile rotates about its base, resulting in translational motion of the end-effector centre in the \(XY\)-plane. Consequently, the predefined regions can become misaligned with the actual end-effector locations during large tilts. Although this misalignment did not produce significant performance degradation in the tested manipulation tasks, it may constrain the accuracy of fine motor control. Incorporating a dynamic, pose-dependent region model that accounts for end-effector displacement could improve precision and robustness.

\section{Conclusion}

This work presents a novel actuator array utilizing an inter-connective flexible material, forming a continuous manipulation surface. This enables the manipulation of smaller objects with a reduced actuator density compared to conventional distributed manipulator systems (\ac{DMS}). The system architecture is scalable to arrays of arbitrary size and supports object translation through two distinct modes of transport, demonstrated across a variety of object geometries. This opens up the opportunity to explore scalable, low-density, distributed manipulation systems well suited to material transport, object positioning, and adaptive handling applications.

\section{Acknowledgments}
This work was conducted as part of the MOZART project, funded by the European Union, EU project id: 101069536.

\bibliography{references}
\bibliographystyle{IEEEtran}
\end{document}